\documentclass[11pt]{article}
\usepackage{coling2020}
\usepackage{times}
\usepackage{url}
\usepackage{latexsym}
\usepackage{amsmath}
\usepackage{booktabs}
\usepackage{multirow}
\usepackage{graphicx}
\usepackage{color}
\usepackage{booktabs}
\usepackage{multibib}
\usepackage{bm}
\usepackage{amsfonts,amssymb}

\usepackage[symbol]{footmisc}

\colingfinalcopy

\title{Cross-lingual Machine Reading Comprehension with Language Branch Knowledge Distillation}

\author{
  \textbf{Junhao Liu}$^{1,2}$\footnotemark[1] \quad \textbf{Linjun Shou}$^3$\footnotemark[1] \quad \textbf{Jian Pei}$^4$ \quad \textbf{Ming Gong}$^3$ \quad
  \textbf{Min Yang}$^1$\footnotemark[2] \quad \textbf{Daxin Jiang}$^3$\footnotemark[2]\\
  $^1${Shenzhen Key Laboratory for High Performance Data Mining,} \\ {Shenzhen Institutes of Advanced Technology, Chinese Academy of Sciences} \\
  $^2${University of Chinese Academy of Sciences} \\
  $^3${Microsoft STCA NLP Group, Beijing, China} \\
  $^4${School of Computing Science, Simon Fraser University} \\
  \small{\{jh.liu,min.yang\}@siat.ac.cn},
  \small{\{lisho,migon,djiang\}@microsoft.com},
  \small{jpei@cs.sfu.ca} \\
}

\date{}

\begin{document}

\maketitle

\footnotetext[1]{Equal contribution. Work was done when Junhao Liu was an intern at Microsoft STCA.}
\footnotetext[2]{Min Yang and Daxin Jiang are corresponding authors.}

\begin{abstract}
Cross-lingual Machine Reading Comprehension (CLMRC) remains a challenging problem due to the lack of large-scale annotated datasets in low-source languages, such as Arabic, Hindi, and Vietnamese. Many previous approaches use translation data by translating from a rich-source language, such as English, to low-source languages as auxiliary supervision. However, how to effectively leverage translation data and reduce the impact of noise introduced by translation remains onerous. In this paper, we tackle this challenge and enhance the cross-lingual transferring performance by a novel augmentation approach named Language Branch Machine Reading Comprehension (LBMRC). A language branch is a group of passages in one single language paired with questions in all target languages. We train multiple machine reading comprehension (MRC) models proficient in individual language based on LBMRC. Then, we devise a multilingual distillation approach to amalgamate knowledge from multiple language branch models to a single model for all target languages. Combining the LBMRC and multilingual distillation can be more robust to the data noises, therefore, improving the model's cross-lingual ability. Meanwhile, the produced single multilingual model is applicable to all target languages, which saves the cost of training, inference, and maintenance for multiple models.
Extensive experiments on two CLMRC benchmarks clearly show the effectiveness of our proposed method.
\end{abstract}

\section{Introduction}
\label{intro}

\blfootnote{
    %
    %
    %
    %
    %
    \hspace{-0.65cm}  
    This work is licensed under a Creative Commons 
    Attribution 4.0 International License.
    License details:
    \url{http://creativecommons.org/licenses/by/4.0/}.
}

Machine Reading Comprehension (MRC) is a central task in natural language understanding (NLU) with many applications, such as information retrieval and dialogue generation. Given a query and a text paragraph, MRC extracts the span of the correct answer from the paragraph. Recently, as a series of large-scale annotated datasets become available, such as SQuAD~\cite{rajpurkar2016squad} and TriviaQA~\cite{joshi2017triviaqa}, the performance of MRC systems has been improved dramatically~\cite{xiong2017dcn+,hu2017reinforced,yu2018qanet,wang2016multi,seo2016bidirectional}. Nevertheless, those large-scale, high-quality annotated datasets often only exist in rich-resource languages, such as English, French and German. Correspondingly, the improvement of MRC quality can only benefit those rich-source languages. Annotating a large MRC dataset with high quality for every language is very costly and may even be infeasible~\cite{he2017dureader}. MRC in low-resource languages still suffers from the lack of large amounts of high-quality training data. 

Besides, in real business scenarios, it is not practical to train separate MRC models for each language given that there are thousands of languages existed in the world. Thus multi-lingual MRC (single model for multiple languages) is of strong practical value by greatly reducing the model training, serving and maintenance costs.


To tackle the challenges of MRC in low-resource languages, cross-lingual MRC (CLMRC) is proposed, where translation systems are used to translate datasets from rich-source languages to enrich training data for low-resource languages~\cite{asai2018multilingual,lee2018semi}. However, CLMRC is severely restricted by translation quality~\cite{cui2019cross}.

Recently, large-scale pre-trained language models (PLM) ~\cite{devlin2018bert,yang2019xlnet,sun2019ernie} are shown effective in NLU related tasks. Inspired by the success of PLM, multilingual PLM~\cite{lample2019cross,Huang2019UnicoderAU,Liang2020XGLUEAN} are developed by leveraging large-scale multilingual corpuses for cross-lingual pre-training. Those powerful multilingual PLM are capable of zero-shot or few-shot learning~\cite{conneau2018xnli,castellucci2019multi}, and are effective to transfer from rich-resource languages to low-resource languages. Although those methods gain significant improvements on sentence-level tasks, such as sentence classification~\cite{conneau2018xnli}, there is still a big gap between the performance of CLMRC in rich-resource languages and that in low-resource languages, since CLMRC requires high quality fine-grained representation at the phase-level~\cite{Yuan2020EnhancingAB}. 

Several studies combine multilingual PLM with translation data to improve the CLMRC performance by either data augmentation using translation~\cite{singh2019xlda} or auxiliary tasks~\cite{Yuan2020EnhancingAB} (see Section~\ref{sec:related} for some details). Those studies take two alternative approaches.  First, they may just leverage translated data in target languages as new training data to directly train target language models \cite{hsu2019zero}. The performance of such models is still limited by the translation issues (i.e, the noise introduced by the translation processing). Second, they may strongly rely on language-specific external corpuses, which are not widely or easily accessible \cite{Yuan2020EnhancingAB}.

According to the generalized cross-lingual transfer result~\cite{lewis2019mlqa}, the best cross-lingual performance is often constrained by the passage language, rather than the question language. In other words, the passage language plays an important role in CLMRC. The intuition is that the goal of MRC to pinpoint the exact answer boundary in passage, thus the language of passage has stronger influence on the performance than the question language. Motivated by this intuition, in this paper, we propose a new cross-lingual training approach based on knowledge distillation for CLMRC. We group the translated dataset (i.e., both questions and passages are translated into all target languages) into several groups. A group, called a \emph{language branch}, contains all passages in one single language paired with questions in all target languages. For each language branch, a separate teacher model is trained. Those language branch specific models are taken as teacher models to jointly distill a single multilingual student model using a novel multilingual distillation framework. With this framework, our method can amalgamate multiple language diversity knowledge from language branch specific models to a single multilingual model and can be more robust to defeat the noises in the translated dataset, which obtains better cross-lingual performance.

We make the following technical contributions.  First, on top of translation, we propose a novel language branch training approach by training several language specific models as teachers to provide fine-grained supervisions.  Second, based on those teacher models, we propose a novel multilingual multi-teacher distillation framework to transfer the capabilities of the language teacher models to a unified CLMRC model.  Last, we conduct extensive experiments on two popular CLMRC benchmark datasets in 9 languages under both translation and zero-shot conditions. Our model achieves state-of-the-art results on all languages for both datasets without using any external large-scale corpus. 

The rest of the paper is organized as follows.  We review related work in Section~\ref{sec:related}, and present our method in Section~\ref{sec:method}. We report experimental results in Section~\ref{sec:exp} and conclude the paper in Section~\ref{sec:con}.

\begin{figure}[t]
    \centering
    \includegraphics[width=0.9\textwidth]{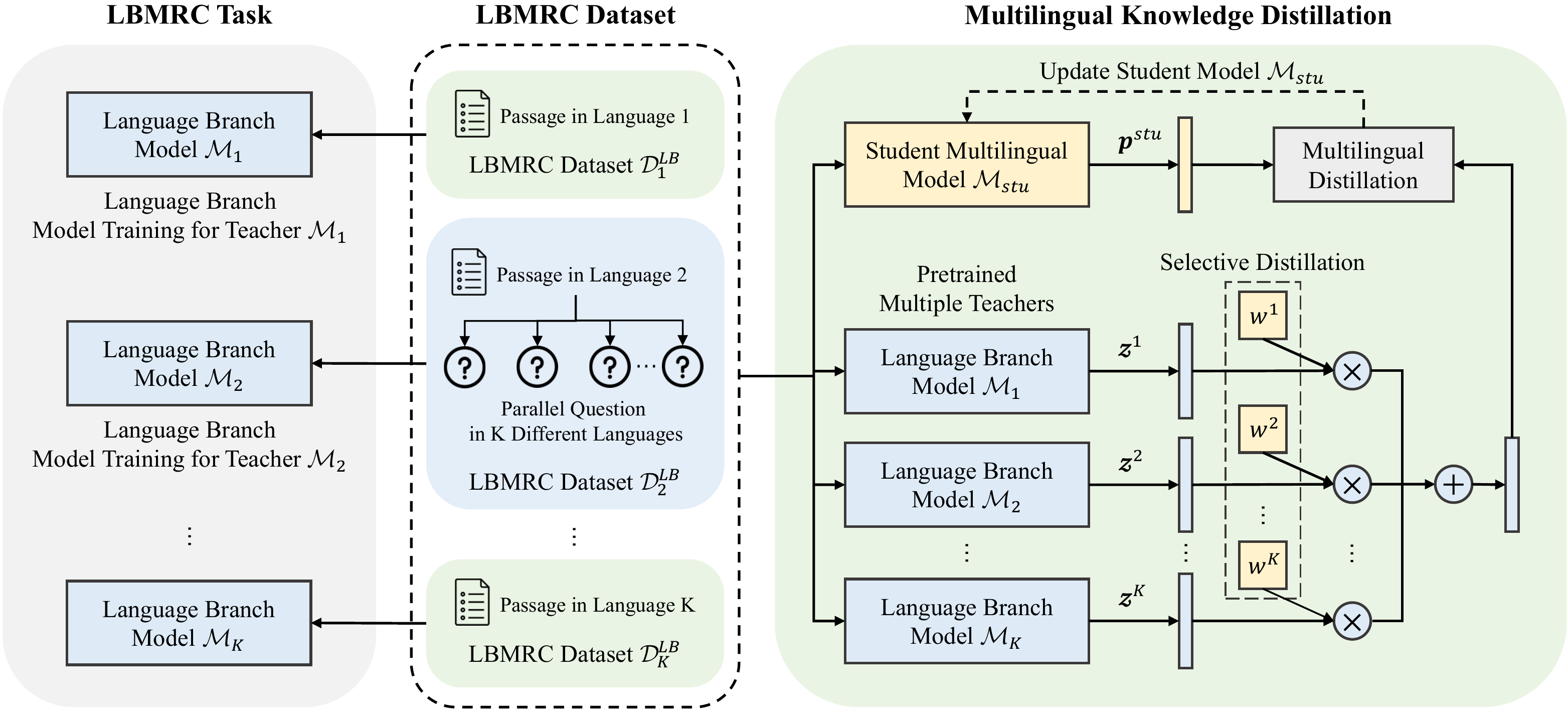}
    \caption{Model overview of our approach.}
    \label{fig:model}
\end{figure}

\section{Related Work}\label{sec:related}

Our study is mostly related to the existing work on CLMRC and knowledge distillation. We briefly review some most related studies here. 

Assuming only annotated data in another source language is available, CLMRC reads one context passage in a target language and extracts the span of an answer to a given question. Translation based approaches use a translation system to translate labeled data in a source language to a low-resource target language. Based on the translated data, \newcite{asai2018multilingual} devise a run-time neural machine translation based multilingual extracted question answering method. \newcite{singh2019xlda} propose a data augmentation strategy for cross-lingual training. \newcite{cui2019cross} leverage a back-translation method to conduct CLMRC. All of these methods rely on a translation system to obtain high-quality translation data, which may not be available for some low-resource languages.

Recently, large-scale pre-trained language models have shown effective in many natural language processing tasks, which prompt the development of multilingual language models, such as multilingual BERT~\cite{devlin2018bert}, XLM~\cite{lample2019cross}, and Unicoder~\cite{huang2019unicoder}. These language models aim to learn language agnostic contextual representations by leveraging large-scale monolingual and parallel corpuses, which show great potential on cross-lingual tasks, such as sentence classification tasks~\cite{hsu2019zero,pires2019multilingual,conneau2018xnli}. However, there is still a big gap between the performance of CLMRC in rich-resource languages and that in low-resource languages, since CLMRC requires the capability of fine-grained representation at the phase-level~\cite{Yuan2020EnhancingAB}. 

To further boost the performance of multilingual PLM on CLMRC task, \newcite{Yuan2020EnhancingAB} propose two auxiliary tasks mixMRC and LAKM on top of multilingual PLM. Those auxiliary tasks improve the answer boundary detection quality in low-resource languages. mixMRC first uses a translation system to translate the English training data into other languages and then constructs an augmented dataset of pairs $\langle$question, passage$\rangle$ in different languages. This new dataset turns out to be quite effective and can be used directly to train models on target languages. LAKM leverages language-specific meaningful phrases from external sources, such as entities mined from search logs of commercial search engines. LAKM conducts a new knowledge masking task. Any phrases contained in the training instances belonging to the external sources are replaced by a special token $[MASK]$.  Then, the task of mask language model~\cite{devlin2018bert} is conducted.  The mixMRC task may still be limited by the translation quality and LAKM requires a large amount of external corpus, which is not easily accessible.

Knowledge Distillation is initially adopted for model compression~\cite{bucilua2006model}, where a small and light-weight student model learns to mimic the output distribution of a large teacher model. Recently, knowledge distillation has been widely applied to many tasks, such as person re-identification~\cite{wu2019distilled}, item recommendation~\cite{tang2018ranking}, and neural machine translation~\cite{tan2019multilingual,zhou2019understanding,sun2020knowledge}. Knowledge distillation from multiple teachers is also proposed~\cite{you2017learning,Yang2020ModelCW}, where the relative dissimilarity of feature maps generated from diverse teacher models can provide more appropriate guidance in student model training. Knowledge distillation is effective in transfer learning in those applications. 

In this paper, on top of translation, we propose a novel approach of language branch training to obtain several language-specific teacher models. We further propose a novel multilingual multi-teacher distillation framework. In contrast to the previous work~\cite{hu2018attention,Yuan2020EnhancingAB}, our proposed method can greatly reduce the noise introduced by translation systems without relying on external large-scale, language-specific corpus. Our method is applicable to more cross-lingual tasks.

\section{Methodology}\label{sec:method}

We formulate the CLMRC problem as follows. Given a labeled MRC dataset $\mathcal{D}_{src}=\{(p^{src}, q^{src}, a^{src})\}$ in a rich-resource language $src$, where $p^{src}, q^{src}$ and $a^{src}$ are a passage, a question, and an answer to $q^{src}$, respectively, the goal is to train a MRC model $\mathcal{M}$ for the rich-resource language $src$ and another low-resource language $tgt$. For an input passage $p^{tgt}$ and question $q^{tgt}$ in $tgt$, $\mathcal{M}$ can predict the answer span $a^{tgt}=(a_{s}^{tgt}, a_{e}^{tgt})$, where $a_{s}^{tgt}$ and $a_{e}^{tgt}$ are the starting and ending indexes of the answer location in passage $p^{tgt}$, respectively.  Model $\mathcal{M}$ is expected to have good performance not only in the rich-resource language $src$, but also in the low-resource language $tgt$.

We first propose a new data augmentation based training strategy, Language Branch Machine Reading Comprehension (LBMRC), to train separate models for each language branch. A language branch is a group that contains passages in one single language accompanied with questions in all target languages. Under this setting, we can construct a language branch dataset for each language. Using each language branch dataset, we train a separate MRC model proficient in the  language. Then, the branch-specific MRC models are taken as multiple teacher models to train a single multilingual MRC student model using a novel multilingual language branch knowledge distillation framework. The overview of our approach is illustrated in Figure~\ref{fig:model}. 

\begin{figure}[t]
    \centering
    \includegraphics[width=0.9\textwidth]{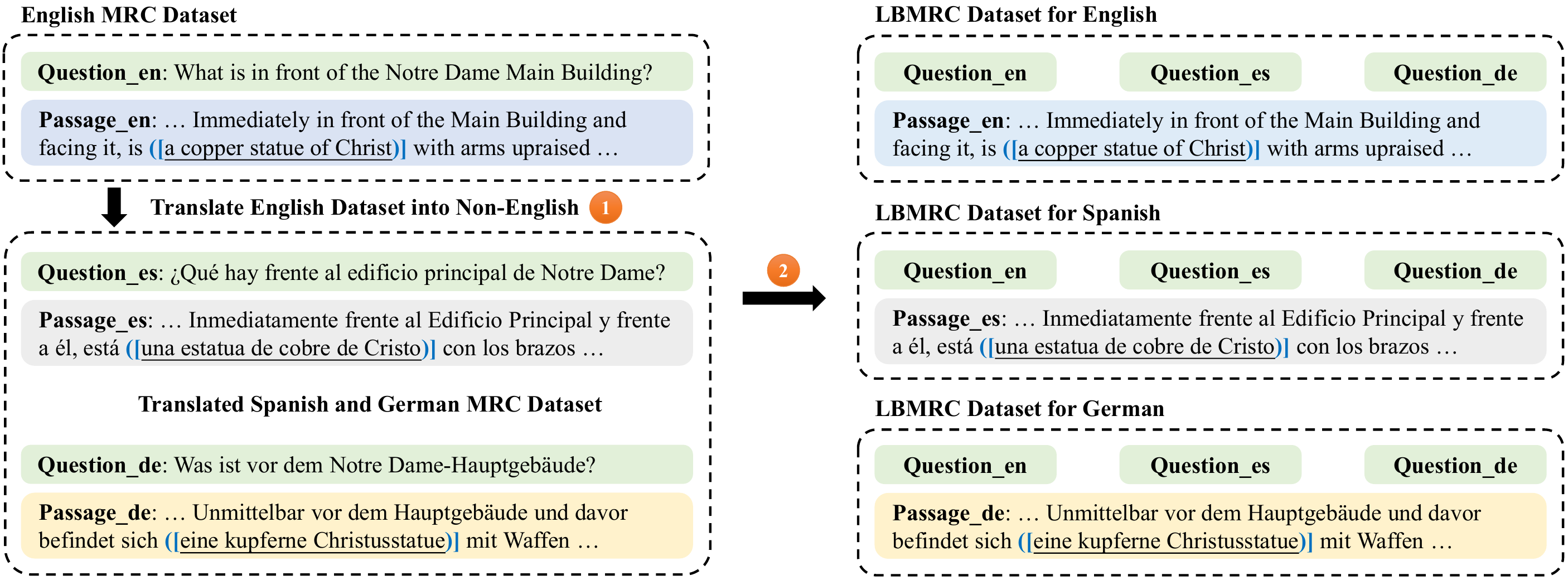}
    \caption{Overview of LBMRC dataset construction process. We use 3 languages (\textit{English}, \textit{Spanish}, \textit{German}) in this illustration. In the first step, the English MRC dataset is translated into the other languages, including both questions and passages. In the second step, the construction method described in Section~\ref{sec:lbmrc_dataset} is applied to build the LBMRC dataset for each language.}
    \label{fig:data}
    \vskip -0.6em
\end{figure}

\subsection{Language Branch Machine Reading Comprehension (LBMRC)}
\label{sec:lbmrc_dataset}

The generalized cross-lingual transfer (G-XLT) approach~\cite{lewis2019mlqa} trains a cross-lingual MRC model using the SQuAD~\cite{rajpurkar2016squad} dataset and evaluates the model on samples of questions and passages in different languages. The results show that the best cross-lingual answering performance in the testing phase is sensitive to the language of passages in the test data rather than the language of questions. This observation suggests that the language of passages in training data may play an important role in the CLMRC task. 

Based on the above understanding, we devise a new data augmentation based training strategy LBMRC. It first trains MRC models in several languages and then distills those models to derive a final MRC model for all target languages.  In contrast to the mixMRC strategy~\cite{Yuan2020EnhancingAB}, LBMRC groups the translation data into several language branches using passage languages as identifiers.  Each language branch contains all passages translated into one single language accompanied with questions in different languages. Figure~\ref{fig:data} shows the overall procedure of this data construction process. We train a separate MRC model for each language branch, which is expected to be proficient in one specific language. 

\subsubsection{Language Branch Construction}

To obtain parallel question and passage pairs in different languages, we adopt a method similar to~\cite{Yuan2020EnhancingAB,singh2019xlda} by employing a machine translation system to translate a labeled dataset of questions and passages in English into datasets in multiple languages $\mathcal{D}_k = \{(p^k, q^k, a^k)\}$, where $p^k, q^k$ and $a^k$ are a passage, a question, and the answer to $q^k$, respectively, all in language $k$. In this process, it is hard to recover the correct answer spans in translated passages. To mitigate this problem, we take a method similar to~\cite{lee2018semi} that adds a pair of special tokens to denote the correct answer in the original passage. We discard those samples where the answer spans cannot be recovered. The language branch for language $k$ is the set of passages and answers in language $k$ accompanied by the queries in all languages, that is, $\mathcal{D}^{LB}_k = \{(p^k, \{q^1, \ldots, q^K\}, a^k)\}$, where $K$ is the total number of languages.

\subsubsection{Language Branch Model Training}
\label{sec:lbmrc}
Similar to the MRC training method proposed in BERT \cite{devlin2018bert}, the PLM model is adopted for encoding the input text $x = [q,p]$ into a deep contextualized representation $\bm{H}\in \mathbb{R}^{L \times h}$, where $L$ represents the length of the input text $x$, $h$ is the hidden size of the PLM model. Then, we can calculate the final start and end position predictions $\bm{p}_s,\bm{p}_e$. Take the start position $\bm{p}_s$ as an example, it can be obtained by the following equations:
\begin{equation}
    \begin{split}
        \bm{z}_s &= \bm{H} \cdot \bm{u}_s + \bm{b}_s \\
        \bm{p}_s &= softmax(\bm{z}_s / \tau)
    \end{split}
    \label{eq:p}
\end{equation}
where $\bm{u}_s \in \mathbb{R}^h, \bm{b}_s \in \mathbb{R}^L$ are two trainable parameters, $\bm{z}_s \in \mathbb{R}^L$ represents the output logits, $\bm{p}_s \in \mathbb{R}^L$ is the predicted output distribution of the start positions, $\tau$ is the temperature introduced by ~\cite{hinton2015distilling} to control the smoothness of the output distribution. 

For each $\mathcal{D}^{LB}_k$, we train a language branch MRC model $\mathcal{M}_k$ by optimizing the log-likelihood loss function:
\begin{equation}
    \mathcal{L}_{NLL}(\mathcal{D}^{LB}_k;\mathcal{M}_k) = -\frac{1}{N}\sum_{i=1}^N \Big( (\bm{a}^k_{s,i})^T \cdot \log(\bm{p}^k_{s,i}) + (\bm{a}^k_{e,i})^T \cdot \log(\bm{p}^k_{e,i}) \Big)
\end{equation}
where $N$ is the total number of samples in $\mathcal{D}^{LB}_k$, the temperature parameter $\tau$ is set to 1, $(\bm{p}^k_{s,i}, \bm{p}^k_{e,i}) \in \mathbb{R}^L$ are the start and end position predictions of sample $i$ from model $\mathcal{M}_k$, $(\bm{a}^k_{s,i}, \bm{a}^k_{e,i}) \in \mathbb{R}^L$ are the ground-truth one-hot labels for the start and end positions of sample $i$ in $\mathcal{D}^{LB}_k$.

\subsection{Multilingual Multi-teacher Distillation}

Let $\mathcal{M}_{stu}$ denote the model parameters of the student multilingual MRC model. $\mathcal{M}_{stu}$ is expected to distill the language-specific knowledge from the multiple language branch teachers $\{\mathcal{M}_k\}_{k=1}^K$. In terms of training data, we take the union of LBMRC datasets as the distillation training dataset $\mathcal{D}$ which is $\mathcal{D} = \{\mathcal{D}^{LB}_1\cup \dots \cup \mathcal{D}^{LB}_K\}$. 

\paragraph{Distillation Training}
We train a multilingual student model to mimic the output distribution of the language branch teacher models. Specifically, the distillation loss of the student model can be described in the form of cross-entropy. In order to distill knowledge from multiple teachers simultaneously, we propose to aggregate the predicted logits from different teachers. Formally, the distillation soft logits $\bm{z}_{s,i}, \bm{z}_{e,i}$ used to train the student model can be formulated as:
\begin{equation}
    \bm{z}_{s,i} = \sum_{k=1}^K w_s^k \cdot \bm{z}_{s,i}^k; \quad
    \bm{z}_{e,i} = \sum_{k=1}^K w_e^k \cdot \bm{z}_{e,i}^k
\end{equation}
where $\textit{w}^k=\{w_s^k, w_e^k\}$ are hyper parameters to control the contributions of each teacher model, $\bm{z}_{s,i}^k$ and $\bm{z}_{e,i}^k$ are the predicted soft logits of sample $i$ from the language branch teacher model $\mathcal{M}_k$. The multilingual multi-teacher distillation loss can be calculated as:
\begin{equation}
    \mathcal{L}_{KD}(\mathcal{D};\mathcal{M}_{stu}) = -\frac{1}{N}\sum_{i=1}^N \Big((\bm{p}_{s,i})^T \cdot \log(\bm{p}_{s,i}^{stu}) + (\bm{p}_{e,i})^T \cdot \log(\bm{p}_{e,i}^{stu}) \Big) \times {\tau}^2
\end{equation}
where $\tau$ is the temperature parameter, $\bm{p}_{s,i}$ and $\bm{p}_{e,i}$ are the start and end distributions calculated by Equation~\ref{eq:p} based on soft logits $\bm{z}_{s,i}, \bm{z}_{e,i}$, $\bm{p}_{s,i}^{stu}$ and $\bm{p}_{e,i}^{stu}$ are also calculated by the softmax-temperature based on the student predicted soft logits $\bm{z}_{s,i}^{stu},\bm{z}_{e,i}^{stu}$. 

Besides, the student model can be also trained using the ground-truth labels of start and end indexes. Let $\mathcal{L}_{NLL}(\mathcal{D};\mathcal{M}_{stu})$ denote the log-likelihood loss function of the one-hot label on the training dataset $\mathcal{D}$, which can be formulated as follows:
\begin{equation}
    \mathcal{L}_{NLL}(\mathcal{D};\mathcal{M}_{stu}) = -\frac{1}{N}\sum_{i=1}^N \Big((\bm{a}_{s,i})^T \cdot \log(\bm{p}_{s,i}^{stu}) + (\bm{a}_{e,i})^T \cdot \log(\bm{p}_{e,i}^{stu}) \Big).
\end{equation}
Finally, the whole multilingual distillation training loss for the student model $\mathcal{M}_{stu}$ can be summarized as:
\begin{equation}
    \mathcal{L}_{Total}(D;\mathcal{M}_{stu}) = \lambda_1 \mathcal{L}_{NLL}(\mathcal{D};\mathcal{M}_{stu}) + \lambda_2 \mathcal{L}_{KD}(\mathcal{D};\mathcal{M}_{stu})
\end{equation}
where $\lambda_1$ and $\lambda_2$ are hyper parameters to balance the contribution of two types of loss.

\paragraph{Selective Distillation}
Here, we consider a proper mechanism to choose the distillation weights $\{\textit{w}^k\}_{k=1}^K$ which can assist the student model to learn from a suitable teacher. We investigate two selection strategies and experiment with their performance in the distillation processing. As the first method, we treat the weights as prior hyper parameters which means that we fix the $\{\textit{w}^k\}_{k=1}^K$ with initial values and train our student model with the same weights during the whole process. In the second mechanism, we use the entropy impurity to measure the teacher’s confidence of a predicted answer including the output distributions of start and end indexes. The confidence of the answer is higher when the impurity has a lower value. Take the start position aggregation as an example, the impurity value is used to determine the weight distribution $\{w_s^k\}_{k=1}^K$ as follows:
\begin{equation}
    \begin{split}
        \bm{p}_s^k &= softmax(\bm{z}_s^k), \ \\
        \mathcal{I}(\bm{p}_s^k) &= - (\bm{p}_s^k)^T \cdot \log(\bm{p}_s^k)  \\
        w_s^k &= \frac{exp(\mathcal{I}(\bm{p}_s^k))}{\sum_j exp(\mathcal{I}(\bm{p}_s^j))}
    \end{split}
\end{equation}
where $\mathcal{I}(\cdot)$ represents the impurity function, $\bm{z}_s^k$ represents the predicted logits of start position from $\mathcal{M}_k$. Based on this, the distillation weights for each teacher model can be adjusted automatically for each instance.

\section{Experiments}
\label{sec:exp}

Extensive experiments of our proposed method are conducted on two public cross-lingual MRC datasets. In the following sections, we describe our experimental settings, results, and analyze the performance.

\subsection{Experimental Settings}

\subsubsection{Datasets and Evaluation Metrics} 

To verify the effectiveness of our method. 
We use the following datasets to conduct our experiments.

\paragraph{MLQA} A cross-lingual machine reading comprehension benchmark~\cite{lewis2019mlqa}. The instances in MLQA cover 7 languages. We evaluate our method on three languages (\textit{English}, \textit{German}, \textit{Spanish}) with translation training method, and also test our method under the setting of zero-shot transfer on the other three languages (\textit{Arabic}, \textit{Hindi}, \textit{Vietnamese}).

\paragraph{XQuAD} Another cross-lingual question answering dataset~\cite{Artetxe:etal:2019}. XQuAD contains instances in 11 languages, and we cover 9 languages in our experiments.
Similar to the setting above, we evaluate our method on \textit{English}, \textit{German}, \textit{Spanish}. In addition, we test our method on \textit{Arabic}, \textit{Hindi}, \textit{Vietnamese}, \textit{Greek}, \textit{Russian}, and \textit{Turkish} under the setting of zero-shot transfer.

\paragraph{Evaluation Metrics} The evaluation metrics used in our experiments are same as the SQuAD dataset~\cite{rajpurkar2016squad} including F1 and Exact Match score. F1 score measures the answer overlap between the predicted and ground-truth answer spans. Exact Match score measures the percentage of predicted answer spans exactly matching the ground-truth labels. We use the official evaluation script provided by~\cite{lewis2019mlqa} to measure performance over different languages. For the XQuAD dataset, we follow the official instruction provided by~\cite{Artetxe:etal:2019} to evaluate our predicted result.

\subsubsection{Baseline Method}

We compare our method with the following baseline methods: (1) Baseline, a method originally proposed in~\cite{lewis2019mlqa} that the MRC model is trained in \textit{English} dataset and tested on the other languages directly, (2) mixMRC, a translation based data augmentation strategy proposed in~\cite{Yuan2020EnhancingAB,singh2019xlda}, which mixes the question and passage in different languages, (3) LAKM, a pre-trained task devised in~\cite{Yuan2020EnhancingAB} by introducing external sources for phrase level mask language model task, and (4) mixMRC + LAKM, a combination method of (2) and (3) through multiple task learning. 


\subsubsection{Implementation Details}

We adopt the pre-trained multilingual language model XLM~\cite{lample2019cross} to conduct our experiments. XLM is a cross-lingual language model pre-trained with monolingual and parallel cross-lingual data to achieve decent transfer performance on cross-lingual tasks. We use the Transformers library from HuggingFace~\cite{wolf2019huggingface} to conduct our experiments. For the MRC task, the pre-trained model is used as the backbone and two trainable vectors are added to locate the start and end positions in the context passage, same with~\cite{devlin2018bert}. 

To construct the LBMRC dataset, We translate the SQuAD dataset to \textit{Spanish} and \textit{German} languages which are two relatively high-resource languages, hence, the number of language branch models is 3 ($K=3$). The target languages of our CLMRC model are \textit{English}, \textit{Spanish}, and \textit{German}. The English branch dataset is always added to other non-English language branch dataset to improve the data quality, which can reduce the impact of noise in data and improve the performance of non-English teachers in our experiments. 

In order to fit the multilingual model into the GPU memory, we pre-processed the teachers' logits for each instance in dataset $\mathcal{D}$.  For the multilingual model training, We use AdamW optimizer with $eps = 1e^{-8}$ and set weight decay to 0.005. The learning rate is set as $1e^{-5}$ for the language branch model training and distillation training. The XLM model is configured with its default setting. For the first selective distillation mechanism, we set the hyper parameters of  $w_s^k = w_e^k = 1/K$ which reach the best performance in our experiments. The distillation loss weight is set as $\lambda_1=0.5, \lambda_2=0.5$ and the softmax temperature $T=2$. We train 10 epochs for each task which can make sure each task converges. 

\subsection{Experimental Results}

\begin{table}[htbp]
\centering
\begin{tabular}{@{}l|cccccc@{}}
\toprule
\multirow{2}{*}{Methods} & \multicolumn{6}{c}{MLQA (EM / F1)}                                                                       \\
                         & en          & es          & de                               & ar          & hi          & vi          \\ \midrule
Lewis\footnotemark[5]                    & 62.4 / 74.9 & 47.8 / 65.2 & \multicolumn{1}{c|}{46.7 / 61.4} & 34.4 / 54.0 & 33.4 / 50.7 & 39.4 / 59.3 \\
Baseline                 & 63.4 / 77.3 & 49.7 / 68.2 & \multicolumn{1}{c|}{48.5 / 63.7} & 37.5 / 56.9 & 37.0 / 54.3 & 42.4 / 63.5 \\
LAKM                     & 64.6 / 79.0 & 52.2 / 70.2 & \multicolumn{1}{c|}{50.6 / 65.4} & - & - & - \\
mixMRC                   & 63.8 / 78.0 & 52.1 / 69.9 & \multicolumn{1}{c|}{49.8 / 64.8} & 38.5 / 58.4 & 40.1 / 57.1 & 45.2 / 66.2 \\
mixMRC + LAKM              & 64.4 / 79.1 & 52.2 / 70.3 & \multicolumn{1}{c|}{51.2 / 66.0} & - & - & - \\ \midrule
Ours-hyper               & \textbf{64.8} / \textbf{79.3} & 53.9 / 71.8 & \multicolumn{1}{c|}{52.1 / 66.8} & \textbf{40.4} / \textbf{60.0} & 42.8 / 59.8 & 46.1 / 67.2 \\
Ours-imp                 & 64.7 / 79.2 & \textbf{54.3} / \textbf{72.0} & \multicolumn{1}{c|}{\textbf{52.4} / \textbf{66.9}} & 40.1 / 59.9 & \textbf{42.9} / \textbf{59.9} & \textbf{46.5} / \textbf{67.5} \\ \bottomrule
\end{tabular}
\caption{EM and F1 score of 6 languages on the MLQA dataset. The left 3 languages (en, es, de) are under translation condition while the right part (ar, hi, vi) results are under the zero-shot transfer method. The results with \footnotemark[5] are adopted from~\newcite{lewis2019mlqa}.}
\label{tab:mlqa}
\vskip -0.5em
\end{table}

\subsubsection{Results on MLQA}

We first evaluate our method on the MLQA dataset in 6 languages. The results are shown in Table \ref{tab:mlqa}. Compared with XLM baselines of original report results and our reproduced results, our method with both selective multilingual distillation strategies (Ours-hyper, Ours-imp) outperform the strong baseline LAKM, mixMRC and mixMRC + LAKM in en, es and de. Especially note that the LAKM method uses extra language corpus to train a better backbone language model, while our method without using external data can also improve the performance significantly with more than 3\% consistent gains in es and de languages. This verifies that the LBMRC training approach could preserve the language characteristics in each teacher model and the multi-teacher distillation step could further reduce the training data noise introduced during the translation process.

We further test our method under the setting of zero-shot transfer in other languages ar, hi, vi. Since the LAKM method requires language-specific corpora to train the backbone model, it is not feasible to access such a corpus to train the backbone model for every low-resource language. Hence, we only compare our method with mixMRC for a fair comparison. For the languages ar, hi and vi, we zero-shot transfer our model to predict in these contexts. We can find that our method also obtains state-of-art results compared with the mixMRC and Baseline with more than 4\% improvement. These results in the MLQA dataset show that our method not only improves the performance in the languages included in our language branch training but also has better-transferring capability to predict the answer in those languages not included in our language branch.  

To compare the two selective distillation strategies we devised above, the impurity selective mechanism Ours-imp gets the best results on most languages, thus proving to be a proper way to aggregate the knowledge from multiple language branch teachers than the weight fixing method Ours-hyper.

\begin{table}[htbp]
\centering
\begin{tabular}{l|ccc}
\toprule
\multirow{2}{*}{Methods} & \multicolumn{3}{c}{XQuAD (EM / F1) }                         \\
                         & en          & es          & de                               \\ \midrule
Baseline                 & 68.8 / 81.3 & 56.9 / 75.6 & \multicolumn{1}{c}{55.5 / 72.6} \\
mixMRC                   & 69.2 / 82.4 & 58.7 / 78.8 & \multicolumn{1}{c}{58.2 / 75.4} \\ \midrule
Ours-hyper               & 69.9 / 83.2 & 59.3 / 79.9 & \multicolumn{1}{c}{59.4 / 76.3} \\
Ours-imp                 & \textbf{70.1} / \textbf{83.4} & \textbf{59.6} / \textbf{80.0} & \multicolumn{1}{c}{\textbf{59.8} / \textbf{76.5}} \\ \bottomrule
\end{tabular}
\caption{EM and F1 score of 3 languages on the XQuAD dataset under the translation-train setting.}
\label{tab:xquad_trans}
\vskip -0.6em
\end{table}

\begin{table}[htbp]
\centering
\begin{tabular}{l|cccccc}
\toprule
\multirow{2}{*}{Methods} & \multicolumn{6}{c}{XQuAD (EM / F1)}                   \\
                         & ar          & hi          & vi          & el & ru & tr \\ \midrule
Baseline                 & 43.2 / 62.6 & 46.0 / 63.1 & 48.7 / 70.4 &  49.4 / 68.5
  &  55.2 / 72.3  &  44.1 / 63.1  \\
mixMRC                   & 42.4 / 63.6 & 50.0 / 66.2 & 52.7 / 72.6 &  51.1 / 72.1  &  58.7 / 75.9  &  47.8 / 65.8  \\ \midrule
Ours-hyper               & \textbf{44.5} / \textbf{65.0} & 52.0 / 67.4 & 55.5 / 74.6 &  52.2 / 73.1  &  59.3 / 76.6  &  \textbf{50.8} / \textbf{68.3}  \\
Ours-imp                 & 44.0 / 64.6 & \textbf{52.5} / \textbf{67.9} & \textbf{55.6} / \textbf{74.9} &  \textbf{52.4} / \textbf{73.3}  &  \textbf{59.6} / \textbf{76.6}  &  50.2 / 67.7  \\ \bottomrule
\end{tabular}
\caption{EM and F1 score of 6 languages on the XQuAD dataset under the zero-shot transfer setting.}
\label{tab:xquad_zero}
\vskip -0.6em
\end{table}


\subsubsection{Results on XQuAD}

We evaluate our method on another common used cross-lingual benchmark XQuAD dataset in 9 languages. The results are shown in Table \ref{tab:xquad_trans} and \ref{tab:xquad_zero} which are under the condition of translation and zero-shot respectively. Since the LAKM method is not suitable in this dataset, we directly compare our method with the mixMRC. Our method consistently outperforms the mixMRC methods in both two conditions. In terms of translation condition, our best method Ours-imp gets 1.7\% and 2.7\% improvement of EM score on es and de respectively. The impurity selective strategy is better for these 3 languages. In terms of the zero-shot transfer, our method obtains a bigger improvement in these 6 languages. Take the Ours-hyper as an example, 4 languages (ar, hi, vi, tr) gain more than 2 points increase of EM score compared with the strong baseline mixMRC. The other 2 languages also have decent EM metric improvement with 2.5\% and 1.5\% for el and ru respectively. The evaluation results on the XQuAD dataset further verify the effectiveness and robustness of our proposed method.


\subsection{Analysis}

\subsubsection{Why Use LBMRC?}
\label{sec:trans}

First, we analyze the training strategy of LBMRC. In order to get different language branch teachers, a possible way is to train each teacher model according to the translated dataset directly. Take the German as an example, this method directly uses the German MRC dataset translated from English as the training set to train the German MRC model. The evaluation results are shown in Table~\ref{tab:trans_train}, where the results with \footnotemark[5] are adopted from~\newcite{lewis2019mlqa}, the Baseline method uses the only \textit{English} dataset to train and then zero-shot transfers to other languages,  and the w/o LBMRC method translates the \textit{English} dataset to other languages and train separate MRC models for each translated dataset. This method (w/o LBMRC) even underperforms the Baseline method where only English training data is used, which aligns with the results from~\newcite{lewis2019mlqa}. The results show that simply leveraging the noisy translated datasets for training can not improve the performance properly.

\begin{table}[h]
\small
    \begin{minipage}{0.49\linewidth}
        \centering
\begin{tabular}{l|ccc}
\toprule
\multirow{2}{*}{Methods} & \multicolumn{3}{c}{MLQA (EM / F1) }                         \\
                         & en          & es          & de                               \\ \midrule
Lewis\footnotemark[5]                 & 62.4 / 74.9 & 49.8 / 68.0 & \multicolumn{1}{c}{47.6 / 62.2} \\
Baseline                 & 63.4 / 77.3 & 49.7 / 68.2 & \multicolumn{1}{c}{48.5 / 63.7} \\
w/o LBMRC                  & - & 49.6 / 68.3 & \multicolumn{1}{c}{47.5 / 62.2} \\ \bottomrule
\end{tabular}
\caption{Translate-train result instead of LBMRC.}
\label{tab:trans_train}
    \end{minipage}
    \begin{minipage}{0.49\linewidth}  
    \centering
    \begin{tabular}{l|ccc}
\toprule
\multirow{2}{*}{Teachers} & \multicolumn{3}{c}{MLQA (EM / F1) }                         \\
                         & en          & es          & de                               \\ \midrule
En                 & 63.2 / 77.6 & 51.2 / 69.4 & \multicolumn{1}{c}{49.3 / 64.2} \\
Es                   & 62.9 / 77.4 & \textbf{51.9} / \textbf{70.0} & \multicolumn{1}{c}{49.1 / 63.9} \\ 
De                 & 63.6 / 78.1 & 51.2 / 69.3 & \multicolumn{1}{c}{\textbf{50.4} / \textbf{65.0}} \\ \bottomrule
\end{tabular}
\caption{Performance of LBMRC teacher models.}
\label{tab:teacher}
    \end{minipage}
\end{table}


The performance of our LBMRC teacher models are shown in Table~\ref{tab:teacher}. With the method introduced in Section~\ref{sec:lbmrc}, we train each language branch teacher model using the according LBMRC dataset. From the results, we can see that the Es and De teacher models achieve the best result on the test set in its own language, which verifies the hypothesis we proposed in Section~\ref{sec:lbmrc} that teacher model trained using LBMRC can preserve language-specific characteristics. Compared with models trained using the mixMRC strategy, LBMRC preserve the language diversity to obtain language-specific expert models.

\subsubsection{Why Multilingual Multi-Teacher Distillation Works?} \label{sec:whywork}

According to Table~\ref{tab:teacher}, an observation is the performance of our teachers is worse than our distillation models (Ours-hyper, Ours-imp) in Table~\ref{tab:mlqa}, which due to the hidden noise in the training set introduced by the translation process. With the help of multilingual distillation training, the student model can be more robust to the data noises and effective to use the translated dataset. We further conduct some ablation studies on different teacher settings: (1) w/o de, remove the de teacher model during multilingual distillation training process, (2) w/o es, remove the es teacher model during multilingual distillation, (3) w/ en, only adopt the en teacher model into the multilingual distillation process, and (4) w/ mix, we take three MRC models trained with mixMRC strategy as the teacher models to do distillation training and obtain a new single student model, where we use the same number of the teacher in our method for a fair comparison. This study (w/ mix) is to verify the effectiveness of the language branch-based multilingual distillation. The ablation results are reported in Table \ref{tab:ablation}.

\begin{table}[htbp]
\centering
\begin{tabular}{l|ccc}
\toprule
\multirow{2}{*}{Exp.} & \multicolumn{3}{c}{MLQA (EM / F1) }                         \\
                         & en          & es          & de                               \\ \midrule
Ours                 & \textbf{64.7} / \textbf{79.2} & \textbf{54.3} / \textbf{72.0} & \multicolumn{1}{c}{\textbf{52.4} / \textbf{66.9}} \\
- w/o de                   & 64.5 / 79.0 & 54.1 / 71.9 & \multicolumn{1}{c}{51.2 / 66.1} \\ 
- w/o es                & 64.4 / 78.6 & 53.2 / 71.2 & \multicolumn{1}{c}{51.8 / 66.6} \\
- w/ en                 & 63.9 / 78.2 & 52.8 / 70.9 & \multicolumn{1}{c}{50.9 / 65.8} \\
- w/ mix                 & 64.6 / 79.1 & 53.6 / 71.3 & \multicolumn{1}{c}{50.8 / 65.5} \\\bottomrule
\end{tabular}
\caption{The ablation study under different experiment settings.}
\label{tab:ablation}
\end{table}

With the ablation study results, we can summarize that each teacher in different languages can have specific contributions to our approach. Take w/o de as an example, the result shows that the de result drops significantly compared with our best score while the en and es results are still relatively close. While the w/o es shows similar trends in terms of the es test result. For w/ en (without leveraging the knowledge from language branch teachers), the results degrade significantly on all languages. To further verify the importance of LBMRC in the multilingual distillation, we replace LBMRC teacher models with models trained using the mixMRC method. The experiment (w/ mix vs Ours) shows that the student model has similar performance in en, but the performance in es and de have a big gap compared with our method, especially for de. This shows that LBMRC could enhance cross-lingual transfer capability and the effectiveness of multilingual distillation.

\section{Conclusion}\label{sec:con}

In this paper, we propose a novel language branch data augmentation based training strategy (LBMRC) and a novel multilingual multi-teacher distillation framework to boost the performance of cross-lingual MRC in low-resource languages. Extensive experiments on two multilingual MRC benchmarks verify the effectiveness of our proposed method either in translation or zero-shot settings. We further analyze the reason why combine the LBMRC and multilingual distillation can gain better cross-lingual performance, which shows that our method is more effective to use the translation dataset and more robust to the noise hidden in the translated data. In addition, our distillation framework produces a single multilingual model applicable to all target languages, which is more practical to deploy multilingual serves.

\section*{Acknowledgments}

Jian Pei's research is supported in part by the NSERC Discovery Grant program. Min Yang's research is partially supported by the National Natural Science Foundation of China (NSFC) (No. 61906185), Guangdong Basic and Applied Basic Research Foundation (No. 2019A1515011705 and No. 2018A030313943), the Youth Innovation Promotion Association of CAS. All opinions, findings, conclusions and recommendations in this paper are those of the authors and do not necessarily reflect the views of the funding agencies.

\bibliographystyle{acl}
\bibliography{coling2020}

\end{document}



\appendix
\section{Implementation Details}
All methods are implemented in PyTorch~\cite{paszke2017automatic} and trained on an Ubuntu 16.04 with 64GB memory and eight GTX 1080 Ti GPU. For all data-sets, we randomly select 80\% of the records as training set, 10\% as validation set and the remaining 10\% as test set. We train the model using training data, and fix model parameters based on the best model performance on validation set. We then test the model on test set. We perform three random runs and report both mean and standard deviation for testing performance. 

We use stochastic gradient descent (SGD) with a learning rate of 2e-5. We use mini-batches of size 64, with batch size 8 for each of 8 GPUs, we use with 1 hidden-layer of 768 hidden units. We use dropout with a rate of 0.5, which is applied to all feedforward neural networks. For the pre-trainng process, We use a batch size of 64 and fine-tune for 4 epochs over the large-scale data-set for two unsupervised task. For each task, we selected the fine-tuning learning rate of 2e-5. For the graph convolutional network, we applied a Bi-LSTM with hidden-layer with 768 hidden units on the top of transformer output. The GCN contains two convolutional layers with the hidden size of 1,536. After node-level convolution, we adapted mean-pooling for graph representation. As to the positional embedding, we created a fixed sinusoidal embedding with 768 hidden units.

For all baseline models, we use pre-trained corresponding transformer models as word embedding and using the output of token [CLS] as sentence embedding. Out-of-vocabulary (OOV) words are hashed to one of 100 random embedding each initialized to mean 0 and standard deviation 1. All other hidden layer weights were initialized from random Gaussian distribution with mean 0 and standard deviation 0.01. Each hyperparameter setting was run on a same machine as the \mname, using Adagrad for optimization with initial accumulator value of 0.1. 

\bibliographystyle{acl}
\bibliography{coling2020}